\documentclass[10pt,journal,letterpaper,compsoc]{IEEEtran}
\pdfoutput=1

\usepackage{graphicx}
\usepackage{subfigure}
\usepackage{placeins}
\usepackage{ragged2e}
\usepackage{rotating}
\usepackage{amsmath}
\usepackage{hyphenat}
\usepackage{floatrow}
\usepackage{amsbsy}
\usepackage{color}
\usepackage{longtable}
\usepackage{multirow}
\usepackage{multicol}

\usepackage[table]{xcolor}
\usepackage{hhline}
\usepackage{booktabs}
\usepackage{floatrow}

\usepackage{mathtools}
\usepackage{amssymb}
\usepackage[pagebackref=true,breaklinks=true,colorlinks,bookmarks=false]{hyperref}
\usepackage{tikz}

\DeclareMathOperator*{\argmax}{argmax}
\newcommand{\ts}[2] {$\underbrace{\text{#1}}_\text{#2}$}

\begin{document}

\title{Looking Beyond Appearances: \\ Synthetic Training Data for Deep CNNs in Re-identification}

\author{Igor~Barros~Barbosa$^{1}$,
        Marco~Cristani$^{2}$,
        Barbara~Caputo$^{3}$,
        Aleksander~Rognhaugen$^{1}$,
        and~Theoharis~Theoharis$^{1}$
\thanks{
$^1$Norwegian University of Science and Technology  (Norway), E-mail: \texttt{igor.barbosa@idi.ntnu.no},
$^2$University of Verona (Italy),
$^3$Sapienza Rome University (Italy).}}

\IEEEtitleabstractindextext{\justify
\begin{abstract}
   Re-identification is generally carried out by encoding the appearance of a subject in terms of outfit, suggesting scenarios where people do not change their attire.
   In this paper we overcome this restriction, by proposing a framework based on a deep convolutional neural network, SOMAnet, that additionally models other discriminative aspects, namely, structural attributes of the human figure (e.g. height, obesity, gender). Our method is unique in many respects. First, SOMAnet is based on the Inception architecture, departing from the usual siamese framework. This spares expensive data preparation (pairing images across cameras) and allows the understanding of what the network learned. Second, and most notably, the training data consists of a synthetic 100K instance dataset, SOMAset, created by photorealistic human body generation software.
   Synthetic data represents a good compromise between realistic imagery, usually not required in re-identification since surveillance cameras capture low-resolution silhouettes, and complete control of the samples, which is useful in order to customize the data w.r.t. the surveillance scenario at-hand, \emph{e.g.} ethnicity. SOMAnet, trained on SOMAset and fine-tuned on recent re-identification benchmarks, outperforms all competitors, matching subjects even with different apparel. The combination of synthetic data with Inception architectures opens up new research avenues in re-identification.

\end{abstract}

}

\maketitle

\IEEEdisplaynontitleabstractindextext

\IEEEpeerreviewmaketitle

\IEEEraisesectionheading{\section{Introduction}\label{sec:introduction}}


\begin{figure*}[!tb]
    \centering
    \includegraphics[width=1\textwidth]{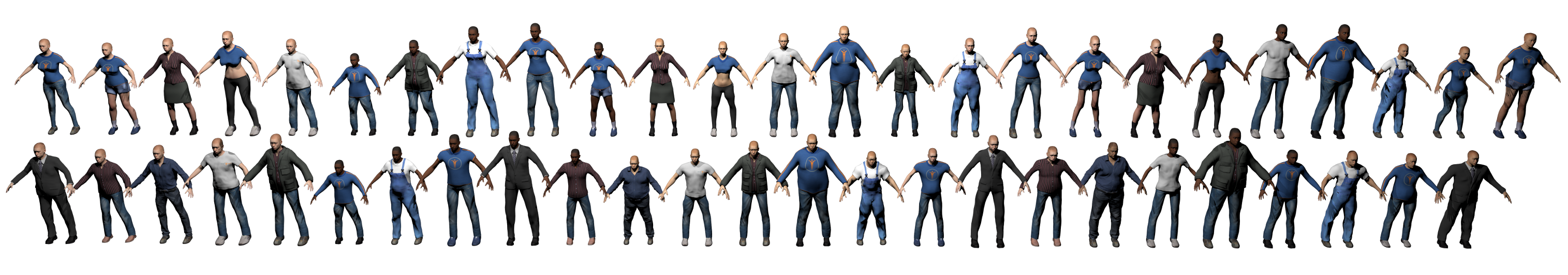}
    \caption{Renderings of the  50  human  prototypes in SOMAset, each one of them wearing one of the 8 sets of clothing available. The top row shows the 25 female subjects and the bottom row the 25  male  subjects.}
    \label{fig:subjects}
\end{figure*}

\begin{tikzpicture}[remember picture,overlay]
\node[anchor=south,yshift=10pt] at (current page.south) {\fbox{\parbox{\dimexpr\textwidth-\fboxsep-\fboxrule\relax}{\textbf{This is the author's version of an article accepted for publication at Computer Vision and Image Understanding 167.
Changes were made to this version by the publisher prior to publication. - https://doi.org/10.1016/j.cviu.2017.12.002}}}};
\end{tikzpicture}%

Re-identification (re-id) aims at matching instances of the same person across non-overlapping camera views in multi-camera surveillance systems~\cite{gong2014person}. Initially a niche application, re-id has attracted huge research interest and has been the focus of thousands of publications in the last five years,
although current solutions are still far from what a human can achieve
\cite{cheng2011custom}.

Recently, deep learning approaches have been customized for re-identification, notably with the so-called \emph{siamese} architectures~\cite{li2014deepreid,Ahmed15, mbdml,GATED,LSTM_REID,personnet}. In a siamese network, a pair of instances is fed into the network, with a positive label when  the instances refer to the same identity, negative otherwise. This causes the network to learn persistent visual aspects that are stable across camera views. An issue with this setting is the setup of the training data: positive and negative pairs should be prepared beforehand, with a significant increase in complexity.

The majority of re-id approaches focuses on modeling the appearance of people in terms of their apparel, with the obvious limitation that changing clothes between camera acquisitions seriously degrades recognition performance.
The RGB-D data provide significantly more information, which explains why there has been considerable progress in this case~\cite{munaro2014one,barbosa2012re}, but, on the other side,  current RGB-D sensors cannot operate at the same distance as typical surveillance cameras; therefore, focusing on RGB does remain an important challenge.

In this paper, we present a re-identification framework based on a convolutional neural network, with the aim of facing the above issues. The framework exhibits several advantageous characteristics.

First, the structure of the network is simpler than a siamese setup. It is based on the Inception architecture~\cite{googlenet}, and is used as a feature extractor. This is similar to a recent approach proposed by \cite{Xiao_2016_CVPR}, which also opted for an Inception-based network architecture. As a by-product, probing inner neurons of deep layers to understand what is learnt by the network is easier than in siamese-like designs. In particular, we show that the network is able to capture structural aspects of the human body, related to the somatotype (gender, being fat or lean, etc.), in addition to clothing information. For this reason, we dubbed the network SOMAnet.

The second unique characteristic of our framework is the data used to train SOMAnet: for the first time we employ a completely synthetic dataset, SOMAset, to train our network from scratch. SOMAset consists of 100K 2D images of 50 human prototypes (25 female and 25 male, Fig. \ref{fig:subjects}), created by mixing three somatotypical ``seeds''~\cite{sheldon1940varieties}:  \emph{ectomorph} (long and lean), \emph{mesomorph} (athletic, small waist) and \emph{endomorph} (soft and round body, large frame), and accounting for different ethnicities. Each of these prototypes wears 11 sets of clothes and assumes 250 different poses, over an outdoor background scene, ray-traced for lifelike illumination. Training networks with synthetic data is not a totally new concept, with pioneering works in 3D object recognition such as~\cite{peng2015learning,zhang2015learning,Borji_2016_CVPR}. However, rendering images of human avatars as a proxy for dealing with real images of people has no precedent in the re-identification literature. We show in our experiments that this choice is effective: when SOMAnet is trained with SOMAset and fine-tuned with other datasets, it achieves state-of-the-art performance on
 four popular benchmarks: CUHK03~\cite{li2014deepreid}, Market-1501~\cite{market}, RAiD~\cite{Das2014} and  RGBD-ID~\cite{barbosa2012re}.

Third, on the RGBD-ID dataset, we are able to show the capability of SOMAnet in recognizing people independently of their clothing, based only on RGB data. This is the first such attempt in the literature, surpassing previous approaches that additionally used depth features. 

The rest of the paper is organized as follows: after reviewing relevant previous work (Sec. \ref{rel-work}), we describe SOMAset (Sec.~\ref{sec:somaset}) and the SOMAnet architecture (Sec.~\ref{sec:somanet}). Sec.~\ref{sec:probe} describes our strategy for probing the inner neurons of the deep layers. Sec.~\ref{Experiments} reports on an exhaustive set of experiments, illustrating the power of the SOMA framework.
Sec.~\ref{Conclusions} concludes with a summary and by sketching future work.

\section{Related literature}
\label{rel-work}

In this section we review the recent literature on  re-identification, focusing in particular on deep learning techniques.  We also discuss the recent trend of creating training sets for recognition, with emphasis on the re-id task.

\subsection{Re-identification}
Strategies for re-identification are many and diverse, with brand new techniques being presented at a vertiginous pace: at the time of writing, Google Scholar gives more than a thousand papers published since January 2016; therefore, it is very hard to recommend an updated survey,  the most recent dating back to 2014 \cite{bedagkar2014survey}.

The early works on re-id were designed to work on single images~\cite{gray2008viewpoint}; batches of multiple images  have been considered afterwards~\cite{farenzena2010person}.
On this input, discriminative signatures are extracted as manually crafted patterns~\cite{farenzena2010person} or
low-dimensional discriminative embeddings \cite{xiong2014person}. While most of the signatures focus on the appearance of the single individuals independently on the camera setting, the study of the inter-camera variations of color (and illumination) gives also convincing results \cite{Martinel_TPAMI,Das2014,Zhang_2015_ICCV,Chakraborty_TPAMI}. Signatures can be matched by exploiting specific similarity metrics \cite{bkak2014re}, which are learned beforehand, thus casting re-id as a metric learning problem \cite{KISSME,Liao_2015_CVPR,Liao_2015_ICCV,Chen_2016_CVPR,Zhang_2016_CVPR}.

Traditionally, re-id assumes that people do not change their clothes between camera acquisitions. The common motivation is that re-id is a short-term operation, thought to cover a time span of  minutes/few hours max, that is, the time necessary for a person to walk between cameras in an indoor environment (an airport, a station etc.). In reality, even in such a time-span, an individual can change his appearance, for example taking off a jacket due to the heating, wearing a backpack etc. Few approaches cover the clothing-change scenario \cite{munaro2014one,barbosa2012re}, all of them relying on RGB-D data. This work is the first that captures structural characteristics of the human figure, in addition to clothing information, exploiting mere RGB data.

The very recent re-id approaches incorporate deep network technology. Typically, they consider image pairs as basic input, where each image comes from a different camera view: when the two images portray the same individual, a positive label is assigned, negative otherwise. These pairs are fed into the so-called \emph{siamese} or \emph{pseudo-siamese} networks, which learn the differences of appearance between camera acquisitions~\cite{li2014deepreid,Ahmed15, mbdml,GATED,LSTM_REID,personnet}. A very recent alternative is the use of triplet loss, where three or more images are compared at the same time~\cite{deep_atrib,chen2017multi}. One image is selected as anchor, while the remaining two images are divided into a positive (having the same identity as the anchor) and a negative one. The objective function over triplets correlates the anchors and the positive images, minimizing their distance. Conversely, the distance from the anchors to the negative images is maximized. Triplet loss has also been used in non-deep learning methods to learn an ensemble of distance functions that can minimize the rank for perfect re-identification \cite{Paisitkriangkrai_2015_CVPR}. One disadvantage of the siamese and triplet loss methods is that they require the dataset to be prearranged in terms of labels. This is cumbersome, may cause highly unbalanced target class distributions and even increase computational complexity.

We argue that re-identification can be carried out by simpler network architectures combined with similarity measurements. This idea was also successfully demonstrated by Sun et al. in 2014 for face verification \cite{Sun2014} where they extract a descriptor using a single path network.

The basis of the proposed approach is to employ a simple (single path) network to learn a descriptor, using a synthetic dataset for training, before fine-tuning on the training partition of a specific dataset. In addition, a probing approach allows the investigation of the characteristics of the network.

\subsection{Training data generation}

In the last years, the design of training sets for recognition has changed from a mostly human-driven operation (crowdsearching data in the most advanced attempts \cite{buhrmester2011amazon}) to a proper research field aiming at automatically producing samples spanning the whole visual semantics of a category, with numbers sufficient to deal with deep learning  requirements~\cite{xia2014well,chen2015webly,cheng2015semantically,baochen14BMVC}.
Two main paradigms do exist: the former assumes that good training data is available on the Internet, and aims at creating retrieval techniques that bridge lexical resources (as \href{https://wordnet.princeton.edu/}{\emph{Wordnet}}) with the visual realm~\cite{xia2014well,chen2015webly,cheng2015semantically}. The latter, most recent, direction assumes that web data is too noisy or insufficient (particularly in the 3D domain) and relies on the generation of photorealistic synthetic data~\cite{baochen14BMVC}. In this case, the trained classifiers should be adapted to the testing situation  by attribute learning \cite{lampert2009learning}, domain adaptation~\cite{glorot2011domain} or transfer learning~\cite{pan2010survey}. This direction seems to be very promising, especially in conjuction with deep architectures~\cite{peng2015learning,zhang2015learning,Borji_2016_CVPR}.

In the re-identification field, the only work that considers the augmentation of a training set by synthetic data is that of~\cite{mclaughlin2015data}, substituting the background scene of the training images with different types of 2D environments. This has been shown to help in reducing the dataset bias, favouring cross-dataset performance. Unfortunately, illumination is not natural in the synthesized samples, and the strategy cannot easily be applied to any dataset (foreground/background segmentation is necessary). Our work goes in the opposite direction, focusing on photorealistic images of the foreground subjects instead of the scenery (which in our case consists in a single, large, outdoor scenario).

\section{The SOMAset dataset}
\label{sec:somaset}

\begin{figure*}[htbp]
    \centering
    \includegraphics[width=0.9\textwidth]{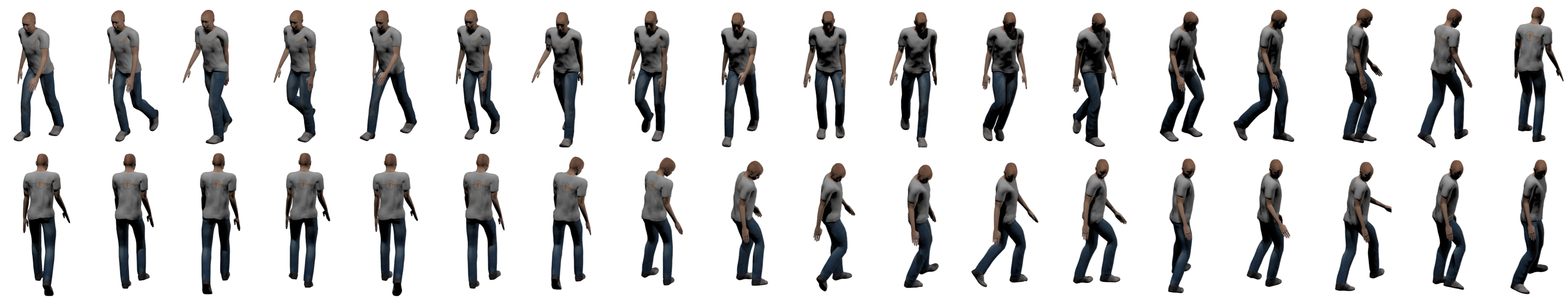}
    \caption{Renderings of a specific subject-clothing assuming 36 out of 250 possible poses. Note the change of the orientation w.r.t. to the camera.}
    \label{fig:poses}
\end{figure*}

In this section we present SOMAset\footnote{SOMAset will be released with a open source license to enable further developments in re-identification.}, describe the protocol followed for creating it and discuss the features that make it
unique compared to other existing re-id collections.

The human figure is normally defined as a mixture of three main somatotypes~\cite{sheldon1940varieties}: \emph{ectomorph} (long and lean), \emph{mesomorph} (athletic, small waist) and \emph{endomorph} (soft and round body, large frame). We account for these facets using an open-source program for 3D photo-realistic human design, \href{http://www.makehuman.org/index.php}{\emph{Makehuman}}, and a rendering engine, \href{https://www.blender.org/}{\emph{Blender}}.
Starting from a generic 3D human model we created 25 male and 25 female subjects, by manually varying the height, weight and body proportions for each subject so as to represent mixtures of the three aforementioned somatotypes.
In order to further improve the similarity to real acquisitions, we also slightly varied parameters like symmetry and the size of legs and/or arms, so as to better simulate natural body variations.

In almost all previous re-identification scenarios, it is assumed that subjects do not change their clothes between camera acquisitions. Re-identification datasets adhere to this assumption, associating identity to appearance (a particular apparel represents a single subject). With SOMAset, we relax this constraint, rendering each of the 50 subjects with 8 different sets of clothing: 5 of these were shared across the sexes while 3 each were exclusive for males / females (thus in total there are 11 types of outfit). In this way, we stimulate the network to focus on morphological cues, other than mere appearance. Experiments with the RGBD-ID dataset (Sec.~\ref{Sec:topexp}) confirm this, having people wearing different clothing between acquisitions.

In more detail, the 3 clothing variations dedicated to females are: T-shirt with shorts; blouse with skirt; sport top with leggings. The 3 male clothing variations are: suit; striped shirt with jeans; shirt with black trousers.

The shared clothing category includes the following 5 variations: white t-shirt with jeans; long sleeve shirt with jeans;  blue T-shirt with jeans; jacket over shirt with jeans; overalls.
Fig. \ref{fig:subjects} shows renderings of the 50 subjects, with female and male subjects in the top and bottom row respectively. The first 8 columns show the 8 clothing possibilities for each gender. To account for ethnicity variations, different skin colors were mapped onto the subjects. Out of the 50 subjects, 16 received Caucasian skin, 16 have darker skin tones, while the remaining 18 have beige skin tones to model Asian types.
We did not include further variations (\emph{e.g.} structural) of the faces and we did omit hair styles, to bound the number of possible variations. Notably, adopting more types of garments does not seem to affect the performance drastically, after some preliminary experiments, not reported here for the lack of space.

Each of the 400 subject-clothing combinations assumed 250 different poses. These poses are extracted from professionally-captured human motion recordings, provided by the CMU Graphics Lab Motion Capture Database \cite{CMUD}. We opted for extracting poses from a recording titled 'navigate', where the subject walks forwards, backwards and sideways. A sample of 36 poses from a specific subject-clothing combination is shown in Fig. \ref{fig:poses}.

Each of the resulting $N=100K$ subject-clothing-pose combinations ($N = 50 ~subjects \times 8 ~clothing~sets \times 250 ~poses$) is placed in a realistic scene (see below) and captured by a virtual camera
with a randomly chosen viewpoint, following a uniform distribution. Specifically, we place the subject in a random location over the floor of the scene, and we take 250 different viewpoints uniformly spanning a hemisphere centered 8 meters away from the subject's initial position. This induces a distance varying from 6 to 10 meters between the camera and the rendered subject-clothing-pose.

\begin{figure}[!htb]
    \centering
    \includegraphics[width=0.8\textwidth]{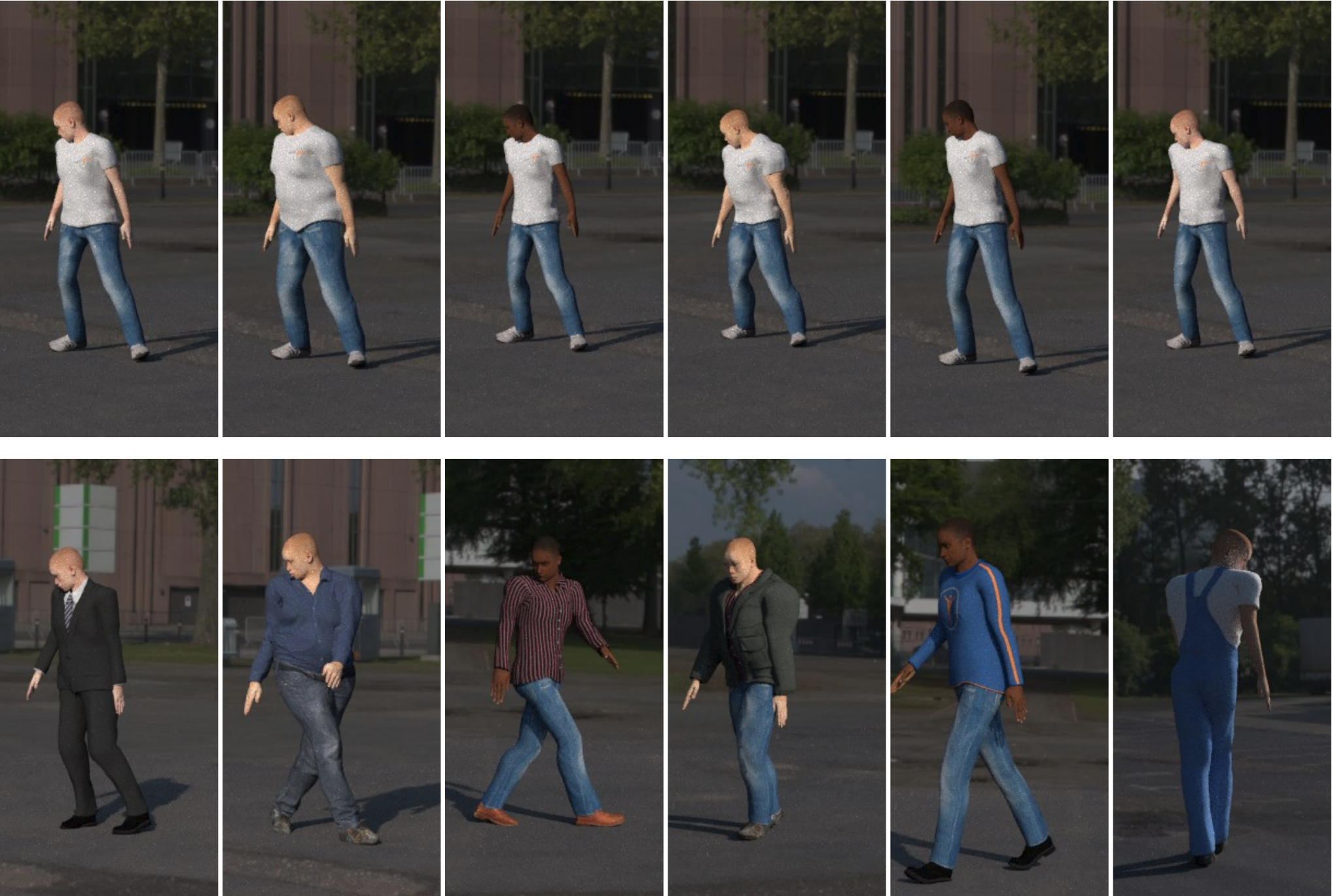}
    \caption{Images sampled from SOMAset. Different male subjects are represented in each column. The second row shows examples where the subjects' pose and clothing vary at the same time. We see that the single 3D environment does yield background variability.}
    \label{fig:collage}
\end{figure}

The different camera viewpoints generate people with diverse image occupancy, different lighting patterns and relative pose w.r.t. the observer. A structured outdoor scene was created for rendering, which covers an area of approximately $900$ m$^2$, where each of the 100K instances was located. The scene includes trees, buildings, pavement, grass and a vehicle, giving a certain variability as the viewpoint changes. A small collage of images from  male subjects of SOMAset is shown in Fig. \ref{fig:collage}.

\vspace{-0.3cm}
\section{The SOMAnet architecture}
\label{sec:somanet}

SOMAnet is a deep neural network that can compute a concise and expressive representation of high level features of an individual, portrayed in an RGB image. This representation enables simple yet effective similarity calculations.

SOMAnet is based on the Inception V3 modules \cite{incep3}, that proved to be well-suited to work on synthetic data  \cite{Modelnet}. Experiments conducted using other frameworks such as Alexnet \cite{Alexnet}, VGG16/VGG19 \cite{Simonyan14c}, Inception V1 \cite{googlenet} and Inception V2 \cite{batch_norm} confirmed this.
The architecture of SOMAnet is described in Sec.~\ref{sec:somanet}; the motivation for our architectural choices
are discussed in Sec.~\ref{sec:DiffGoog} and Sec.~\ref{sec:DiffSiam}. Subsequently, we present the pipeline for training SOMAnet from scratch in Sec.~\ref{Sec:training}, and  the fine-tuning strategy to customize it to diverse testing scenarios, together with the re-identification algorithm, in Sec.~\ref{sec:transfer}.

\subsection{Architecture}

\begin{figure}[!htb]
\centering
   \includegraphics[width=0.9\textwidth]{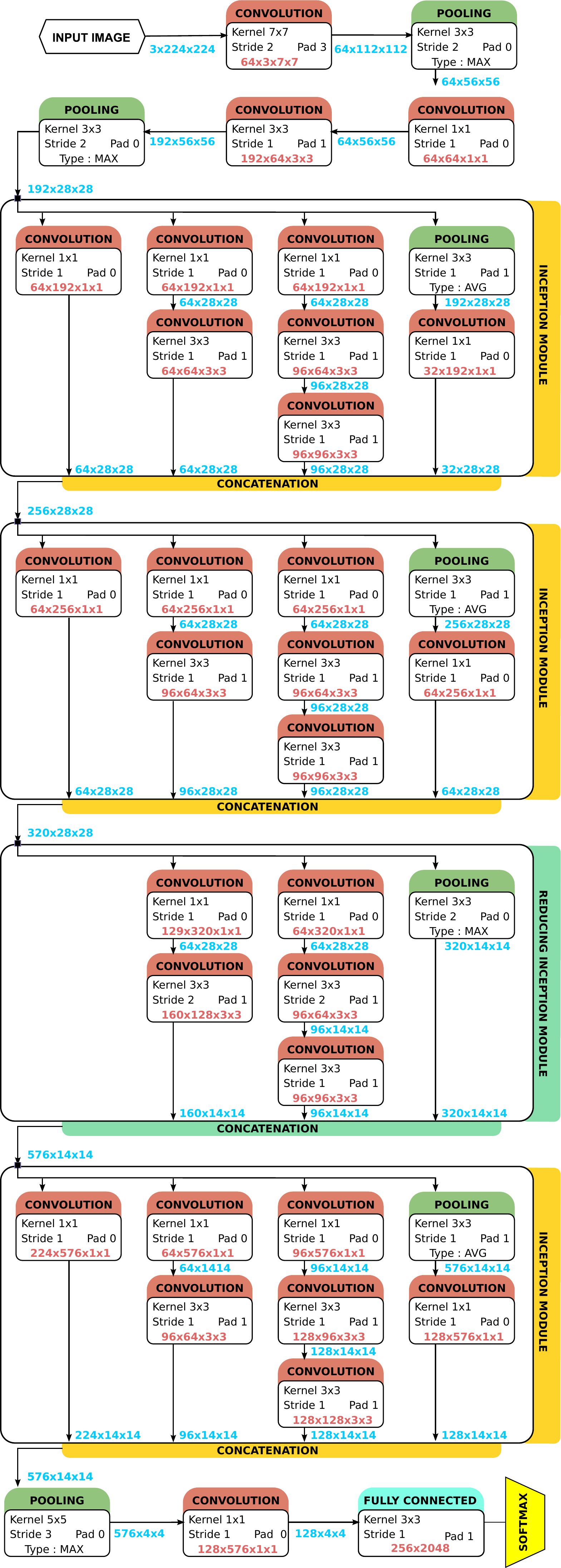}
    \caption{SOMAnet architecture; each layer has its respective activation outputs presented in blue text. }
    \label{fig:final_network}
\end{figure}

Our architecture follows closely the Inception V3 model  \cite{incep3} (Fig.~\ref{fig:final_network}): the initial sequence of convolutions and max pooling replicates  the original architecture. These are followed by two cascading Inception modules and a modified Inception module (Reduced Inception Module) that reduces the input data size by a half by using larger strides in the $3\times3$ convolution and in the pooling layer. Moreover, it drops the $1\times1$ convolution windows that would be used as output of the inception module. The network proceeds to a fourth inception module providing data to our last layers; a max pooling layer followed by a convolution layer that feeds the fully connected layer leading to the output softmax layer.

The use of $3\times3$ windows is preferred over other window sizes, because they are more computationally efficient than larger convolutions used in previous works. A cascade of $3\times3$ convolution windows can provide a proxy for the analysis derived by $5\times5$ and $7\times7$, which were used in \cite{googlenet,batch_norm}. The convolution layers in our network uses rectified activation units (ReLUs) \cite{glorot2011deep} which have sparse activation and efficient gradient propagation as they are less affected by vanishing or exploding gradients. Unlike previous Inception networks \cite{googlenet,batch_norm,incep3} our fully connected layers employ the hyperbolic tangent as activation unit.

 We performed a toy experiment on SOMAset + SOMAnet to sense the complexity of a re-id task where a given synthetic subject (out of the gallery of the 50 different identities) can wear different clothes. The 100,000 images have been partitioned into training ($70\%$ of the total images), validation and testing sets ($15\%$ each); the validation and testing sets were provided so that users of our dataset can assess problems in training, such as overfitting and underfitting. The images have been sampled so to have a proportional number of instances for each subject within each partition. The re-identification results, following the algorithm explained in Sec.~\ref{sec:transfer}, in terms of Cumulative Matching Characteristic (CMC) curve recognition rate at predefined ranks, are reported in Table~\ref{tab:somaset}.
We see that this training strategy allows
 to get an adequate descriptor for the somatotypical characteristics of the subjects, giving $79.69\%$ rank 1 success rate on the testing set.

\begin{table}[!hb]
\scriptsize
\centering
  \begin{tabular}{lrr}
    \toprule
    \multicolumn{3}{c}{SOMAset} \\
    \cmidrule(r){1-3}
    Set             &  Rank 1    & Rank 5 \\
    \midrule
    Validation      & $99.77\%$  & $100.00\%$ \\
    Testing         & $79.69\%$  & $99.75\%$ \\
    \bottomrule
  \end{tabular}
\caption{SOMAnet classification performance on the rendered SOMAset.}
\label{tab:somaset} \vspace{-0.3cm}
\end{table}

\subsubsection{Difference to GoogLeNet}
\label{sec:DiffGoog}

The GoogLeNet inception network~\cite{googlenet, batch_norm, incep3} was designed for the Large Scale Visual Recognition challenge \cite{imagenet}. Hence, it needed to be deep enough to learn abstract features able to differentiate up to a thousand different classes. Such deep architecture might be unnecessary for more specific image recognition tasks.  The original design of GoogLeNet also presents three objective functions,
conceived to help with gradient propagation as the network becomes deeper.

To assess the appropriate depth of the Inception architecture in the case of the re-id task, we used the rendered SOMAset. We designed a classification task, where the original GoogLeNet network must correctly classify all the subjects' identities.
We used the experimental setting described
above.

The original GoogLeNet was trained until the validation set reached a plateau for all its three objective functions. Results showed that, for the specific task of re-id using SOMAset, there was no performance gain by using the deeper classification stages. The network was thus re-designed to use only four Inception V3 modules. Consequently, the network did not need multiple outputs to help in gradient propagation (Fig.~\ref{fig:final_network}).

SOMAnet also differs from previous versions of GoogLeNet in the fully connected layer, where the hyperbolic tangent is used as the activation function, because it is zero-centered and has a bounded output space. The output of the fully connected layer of SOMAnet produces a vector $\mathcal{X} \in \mathbb{R}^{256}$ within $[-1,1]$. This enforces a new embedding computation, with a dimensionality reduction from $2048$ to $256$ dimensions.

\subsubsection{Difference to Siamese networks}
\label{sec:DiffSiam}

Siamese networks have been successfully employed in re-identification by reformulating the task as a binary classification problem \cite{li2014deepreid,Ahmed15}.
Because the input space of siamese networks is expanded from one image to two, complexity challenges arise when training such  networks on large datasets. Space requirements increase as the square of the input images.
Thus, it becomes infeasible to process the complete set of combinations during training time and one needs to select which image pair samples to use in order to have a balanced training set.
In the case of pseudo-siamese networks, training resources must be spent in learning the convolution weights of each different input branch. This effectively makes the network wider in shallow layers.

Our architecture does not suffer from these issues (Fig.~\ref{fig:final_network}). It
 computes a compact latent embedding space (in our case a $\mathbb{R}^{256}$ descriptor)  and we thus use the network as a feature extractor, with linear space and time requirements.
Given the descriptor, similarity distances are setup and evaluated to perform re-id, as explained in Sec.~\ref{sec:transfer} .

\subsection{Training Phase}\label{Sec:training}

SOMAnet is trained using backpropagation \cite{Rumelhart_BP} to minimize the cross-entropy objective function.

Parameters are optimized using a mini-batch gradient descent method with momentum and weight decay \cite{sutskever2013momentum}. This type of training strategy has been shown to be effective \cite{Alexnet, googlenet, Simonyan14c}. Here, 32 images are used per mini-batch. The SOMAnet model uses the Xavier initialization of weights, which is a good starting point for deep neural networks \cite{XAVIER}.

The cross-entropy objective function expects that both the target and predictive outcomes are probability distributions. This constraint can be achieved by encoding target outcomes as one-hot vectors, while the predictive outcomes produced by the neural network can be transformed into a distribution by using the softmax function as seen in Fig.~\ref{fig:final_network}. The learning rate is initialised as $\alpha=0.1$ and is reduced by a factor of $10$ whenever the objective function reaches a plateau. We can successfully use high learning rates because the proposed model uses batch normalization, as presented in \cite{batch_norm}. SOMAnet was trained using the Caffe package~\cite{caffe} on an NVIDIA GeForce GTX TITAN X GPU.
Training the architecture on the full SOMAset took 3 hours.

\subsection{Adapting SOMAnet to Real Data and Re-identification}
\label{sec:transfer}

To deal with testing sets made up of real people, domain adaptation strategies have to be included, that in the case of deep neural networks amounts to fine-tuning \cite{DonahueJVHZTD14}.
Specifically, we force the fine-tuning to focus on the actual classification task (softmax layer). We expect this to help avoid over-fitting of shallower layers, while at the same time giving the chance to obtain strong results with little target data.
We also want to avoid the layer specificity problem \cite{yosinkiFT}. Therefore, we allow fine-tuning to take place in shallower layers but with smaller learning rates. This forces the transferred SOMAnet to comply with the new re-id task by only changing the initial layers a little.
The fine-tuning protocol is summarized here:
\begin{enumerate}
    \item \textbf{Transfer SOMAnet to a new task} by replicating all layers except for the final softmax layer.
    \item \textbf{Set the learning rates} of all layers  preceding  the softmax layer to be ten times smaller than that of the final layer.
\end{enumerate}

After fine-tuning, the output of the penultimate layer
is used as feature descriptor. This vector individuates a $\mathbb{R}^{256}$ latent space which is bounded within $[-1, 1]$ and represents an embedding suitable for efficient distance computations. To add more invariance to the descriptor, we mirror the input image, extract another 256-dimensional feature vector and concatenate it with the original one, obtaining a 512-dimension descriptor.

To compute the distance between descriptor $\mathbb{F}_Q$  and $\mathbb{F}_G$ of a query image $Q$ and a gallery image $G$,
we opted for the cosine distance. In preliminary tests this distance function was shown to be effective. The distance between descriptors gives the output of the re-id, that is, the rank of the gallery images w.r.t. the distance to the query sample.

\section{Experiments}
\label{Experiments}
In this section we explore the potential of SOMAnet and SOMAset, focusing on different aspects of the network and analysing the contribution of the dataset, by performing quantitative and qualitative experiments.
After describing the benchmarks used (Sec.~\ref{Sec:dataset}), we focus on SOMAnet (Sec.~\ref{Sec:SOMAnetEvaluation}), we describe experiments illustrating its performance against other deep architectures (Sec.~\ref{Sec:vsDeep}), we show how some neurons encode specific features of humans (Sec.~\ref{sec:rgbdid_exp}) and how a synthetic training dataset has a positive impact on a deep architecture (Sec.~\ref{Sec:topexp}). Then, we consider SOMAset (Sec.~\ref{sec:SOMAset}), illustrating its role in increasing re-id performance (Sec.\ref{Sec:ScratchTraining}), and exploring how different versions of SOMAset (different number of subjects and poses) change the recognition scores (Sec.~\ref{sec:reduce_sub} and Sec.~\ref{sec:reduce_pos}, respectively).

\subsection{Datasets}\label{Sec:dataset}
We briefly present here the four datasets that we focus on, highlighting the different challenges they represent in terms of re-id.
For comparative purposes, for each dataset we consider state-of-the-art peer-reviewed methods in terms of the Cumulative Matching Characteristic (CMC) curve, and the mean Average Precision (mAP).

\subsubsection{CUHK03}
CUHK03 is the first person re-id dataset large enough for deep learning \cite{li2014deepreid}, with an overall 13164 images. It consists of 1467 identities, taken from five cameras with different acquisition settings. Each identity is observed by at least two disjoint cameras.

The images are obtained from a series of videos recorded over months, thus incorporating drastic illumination changes caused by weather, sun directions, and shadow distributions (even considering a single camera view). As usual in the literature, we consider here the dataset version where pedestrians in the raw images are manually cropped to ease the re-id.

The evaluation of re-id performance on this dataset follows two protocols, one for single-shot and another for multi-shot. In the single-shot case, we follow the protocol of  \cite{li2014deepreid}, commonly adopted in the literature: the dataset is partitioned into a training set of 1367 identities and a test set of 100 identities; during evaluation, we randomly take one of the test set images from each identity of a camera view as probe, using another camera for the corresponding gallery set images (where there exists one image of the same identity as the probe). In the multi-shot case, there are multiple images of each identity in both the probe and gallery sets. We thus compute the average distance from all the probe images w.r.t. all images of the gallery set, producing a ranking.  All the experiments are evaluated with 10 cross validations using random training/test set partitions.

\subsubsection{Market-1501}

Market-1501 is the largest real-image dataset for re-id so far, containing 1501 identities over a set of 32668 images, where each image portrays a single identity \cite{market}.
Five high-resolution and one low-resolution camera were used in the dataset acquisition. Each identity is present in at least two cameras. The dataset is partitioned as follows: the training set consists of 750 identities and 12936 images; these are the images used for training/fine-tuning SOMAnet. The remaining 751 identities are contained in a test set of 19732 images, i.e. 3368 query images which are matched against a gallery set of 16364 images ($19732-3368$).





The testing protocol has been specified in~\cite{market}, and the code for the perfomance evaluation has been provided by the authors. In the single-shot re-id modality, each query image is compared against the gallery images, excluding those that refer to the subject captured by the same camera view (for each query image, there are an average of 14.8 cross-camera ground truths). The mean average precision (mAP) metric is employed, since it is capable of measuring the performance with multiple ground truths.
The dataset also contains extra sets of images for each of the 3368 identities in order to allow testing a multi-shot scenario.

\subsubsection{RAiD}
RAiD (Re-identification Across indoor-outdoor Dataset) is a 4-camera dataset where a limited number of identities (41) is seen in a wide area camera network \cite{Das2014}. The images of RAiD  have  large  illumination  variations  as  they  were  collected  using  both indoor  (cameras  1  and  2)  and  outdoor  cameras  (cameras  3  and  4). The protocol for re-id is the following: the subjects are randomly divided  in two sets, training (21 identities) and testing (20 identities). In total there are 6920 images, for an average of around 161 images per identity.  The training set (around 3300 images in total, depending on the chosen subjects) is used to fine-tune SOMAnet; this represents a challenge due to the small number of data, when compared to the other, more recent, repositories.

For evaluating the multi-shot modality, 10 images for each test identity are picked as query from a single camera and the images associated with a different camera are used as gallery set. Specifically, we evaluate the camera pairs 1-2, 1-3, 1-4, where the latter two configurations have large inter camera illumination variations
Evaluation is done using five cross-validation rounds. For each round, a new random identity partition is made for creating the training and test set, always keeping the proportion of 21/20 subjects for training/testing.

\subsubsection{RGBD-ID}\label{Sec:RGBDdataset}

The RGBD-ID dataset has been originally crafted to explore depth data in a re-id scenario. It contains four different groups of data, all from the same 79 people (identities): 14 female and 65 male. The  first ``Collaborative'' group has been obtained by recording, in an indoor scenario, with a Kinect camera (RGB + depth data), a frontal view of the people, 2 meters away from the camera, walking slowly, avoiding occlusions and with stretched arms. The second group (``Backwards'') consists of back-view acquisitions of the people while walking away from the camera. The third (``Walking1'') and fourth (``Walking2'') groups of data are composed by frontal recordings of the people walking normally while entering a room in front of the camera. There are in average 5 frames of RGBD data per person per group. It is important to note that people in general changed their clothes between the acquisitions related to the four groups of data; most cloth changes occur between groups ``Walking1'' and ``Walking2'' (59 cases out of 79). Additionally, in the ``Walking2'' group 45 out of the 79 people have the same t-shirt, in order to simulate a work environment where people wear the same attire.

In the experiments, we use the ``Collaborative'' and the ``Backwards'' groups for fine-tuning, keeping ``Walking2'' as probe set and ``Walking1'' as gallery set.

Note that we introduce here a new way to use the RGBD-ID data. Previous studies mostly focus only on the depth data to obtain reasonable results, thereby ignoring the RGB imagery. This is because the change of outfit between acquisitions makes the re-id problem harder. Preliminary studies carried out in~\cite{barbosa2012re} reported low performance when using RGB data only. In contrast, we only use the RGB data here to see whether SOMAnet can extract structural aspects of the human silhouette from them.

\subsection{Analysis of SOMAnet}\label{Sec:SOMAnetEvaluation}

\subsubsection{Comparing SOMAnet to other deep architectures}\label{Sec:vsDeep}

The first experiments show the advantage of SOMAnet w.r.t. other deep network architectures, such as siamese-inspired architectures. To this aim, we use the CUHK03 and Market-1501 datasets. We train and test on the respective partitions of each dataset, obtaining \ts{SOMAnet}{CUHK03} and \ts{SOMAnet}{Market-1501}.

Comparative results for the CUHK03 dataset against other recent deep network architectures are reported in Table~\ref{Table:CUHK03singleshot_training}.


\begin{table}[htbp]
\scriptsize
	\centering
\begin{minipage}[b]{0.9\columnwidth}
	\caption{Analysis of the performance of SOMAnet against other deep network architectures when trained and tested exclusively on CUHK03, in the \emph{single-shot} modality} \label{Table:CUHK03singleshot_training}
	\end{minipage}

	\begin{tabular}{cccccc}
		\toprule
		\multirow{ 2}{*}{\textbf{}} & \multicolumn{5}{c}{\textbf{Rank}} \\
		\cmidrule(r){2-6}
		 & \textbf{1} & \textbf{2}& \textbf{5} & \textbf{10} & \textbf{20} \\
		\midrule
    FPNN \cite{li2014deepreid}    & 20.65  & 32.53 & 50.94 & 67.01 & 83.00 \\
    JointRe-id \cite{Ahmed15}     & 54.74  & 70.04 & 86.50 & 93.88 & 98.10 \\
    LSTM-re-id \cite{LSTM_REID}   & 57.30  & -     & 80.10 & 88.20 & -     \\
    Personnet \cite{personnet}    & 64.80  & 73.55 & 89.40 & 94.92 & 98.20 \\
    MB-DML \cite{mbdml}           & 65.04  & -     & -     & -     & -     \\
    Gated \cite{GATED}            & 68.10  & -     & 88.10 & 94.60 & -     \\
    DGD-CNN \cite{Xiao_2016_CVPR} & 72.60 & -     & -     & -     & -     \\
    MTDnet \cite{chen2017multi} & \textbf{74.68}  & - & \textbf{95.99} & \textbf{97.47} & -     \\
    \ts{SOMAnet}{CUHK03} & 68.90 & 82.10  & 91.00  & 95.60 & 98.30 \\
    \bottomrule
    \end{tabular}
\end{table}

The top three performers when trained exclusively on CUHK03 are DGD-CNN, \ts{SOMAnet}{CUHK03} and MTDnet; note that two out of these three networks are Inception-based while the third is a multitask network trained with triplet loss.


With respect to the multi-shot modality, the only approach we can compare to is MB-DML \cite{mbdml,Lin_2015_ICCV}, a siamese architecture based on bilinear convolutional neural networks and deep metric learning. Results are shown in Table~\ref{Table:CUHK03multishot_training}, with \ts{SOMAnet}{CUHK03} achieving the best score.

\begin{table}[htbp]
\scriptsize
	\centering
	\caption{Analysis of the performance of SOMAnet against another deep network architecture when trained and tested exclusively on CUHK03, in the \emph{multi-shot} modality. (Note that only rank-1 performance in the multi-shot modality is provided by the MB-DML paper~\cite{mbdml}).}
	\begin{tabular}{cccccc}
		\toprule
		\multirow{ 2}{*}{\textbf{}} & \multicolumn{5}{c}{\textbf{Rank}} \\
		\cmidrule(r){2-6}
		 & \textbf{1} & \textbf{2}& \textbf{5} & \textbf{10} & \textbf{20} \\
		\midrule
        MB-DML \cite{mbdml} & 80.60   & -   & -   & -   & -       \\
        \ts{SOMAnet}{CUHK03}    & \textbf{83.60}   & \textbf{93.40}   & \textbf{97.50}   & \textbf{99.20}   & \textbf{99.70}   \\
    \bottomrule
    \end{tabular} \label{Table:CUHK03multishot_training}
\end{table}


Single- and multi-shot results on the Market-1501 dataset are shown on Table~\ref{Table:Market1501singleshot_training}. For the singe-shot modality, SOMAnet surpasses the other methods in terms of CMC ranks and mean average precision.

\begin{table}[htbp]
\scriptsize
	\centering
	\caption{Analysis of the performance of SOMAnet against other deep network architectures when trained and tested exclusively on Market-1501, in both \emph{single-shot} and \emph{multi-shot} modalities.}\label{Table:Market1501singleshot_training}
	\scalebox{0.82}{
	\begin{tabular}{cccccccc}
        \toprule
		\multirow{ 2}{*}{\textbf{Single-shot}} & \multicolumn{6}{c}{\textbf{Rank}}  & \multirow{ 2}{*}{\textbf{mAP}} \\
		\cmidrule(r){2-7}
		\textbf{}& \textbf{1} & \textbf{5}& \textbf{10} & \textbf{20} & \textbf{30} & \textbf{50}&  \\
		\midrule
		Personnet \cite{personnet}    & 37.21 & -       & -       & -       & -       & -       &  18.57 \\
		SSDAL \cite{deep_atrib} & 39.40 & -       & -       & -       & -       & -       & 19.60\\
    	MB-DML \cite{mbdml}  & 45.58 & -       & -       & -       & -       & -       & 26.11\\
        Gated \cite{GATED}  & 65.88 & -       & -       & -       & -       & -       & 39.55\\
        \ts{SOMAnet}{Market-1501} & \textbf{70.28} & \textbf{87.53} & \textbf{91.69} & \textbf{94.57} & \textbf{95.64} & \textbf{96.85} & \textbf{45.05}\\
        \midrule
        \multirow{ 2}{*}{\textbf{Multi-shot}} & \multicolumn{6}{c}{\textbf{Rank}}  & \multirow{ 2}{*}{\textbf{mAP}} \\
		\cmidrule(r){2-7}
		\textbf{}& \textbf{1} & \textbf{5}& \textbf{10} & \textbf{20} & \textbf{30} & \textbf{50}&  \\
		\midrule
		SSDAL \cite{deep_atrib} & 49.00 & -       & -       & -       & -       & -       &25.80\\
		MB-DML \cite{mbdml}  & 56.59 & -       & -       & -       & -       & -       & 32.26\\
        LSTM-re-id \cite{LSTM_REID}  & 61.60 & -       & -       & -       & -       & -       & 35.3\\
        Gated \cite{GATED}  & 76.04 & -       & -       & -       & -       & -       & 48.45\\
        \ts{SOMAnet}{Market-1501} & \textbf{77.49} & \textbf{91.81} & \textbf{94.69} & \textbf{96.56} & \textbf{97.27} & \textbf{98.25} & \textbf{53.50} \\
    \bottomrule
    \end{tabular}
    }
\end{table}

These first experiments show that SOMAnet, independently from the training dataset, leads to state-of-the-art re-id results. As we will see in Sec.~\ref{Sec:topexp}, the adoption of SOMAset as training data gives even higher scores.

\subsubsection{Probing specialized neurons in SOMAnet}
\label{sec:rgbdid_exp}
\label{sec:probe}

Several insights have come from attempts at visualising what specific neurons respond to, for a given convolutional neural network
\cite{erhanVZ09,SimonyanVZ13,zeilerVZ14}. Some approaches pose the problem of understanding neuron behavior as an optimization problem, where images are propagated through the network  to find  which image region maximizes the activation of a particular neuron \cite{Girshick_2014_CVPR}. Other visualization techniques have been used to identify neurons that respond to specific visual stimuli; for example, \cite{yosinskiVZ15} individuates a neuron responsive to face patterns in a CNN  trained for the Large Scale Visual Recognition challenge \cite{imagenet}. Although visualization methods can be used for finding specialized neurons, it can be a slow task since the analysis of results is still manual.

Here we propose a different  approach: the goal is to find a specialized neuron $N_S$ over a given set of neurons which gives the highest response to a given stimuli or characteristic (e.g. gender, obesity, a particular type of clothing) and is unlikely to respond to other characteristics. The search  can be formulated as solving the optimization problem:

\begin{equation}
  N_S = \argmax_{N} \mathcal{D}(\mathbf{C},\mathbf{R},N),
   \label{Eq:disc}
\end{equation}

where $\mathcal{D}$ is a \emph{discernibility} measurement between two sets of images $\mathbf{C}$ and $\mathbf{R}$, given a neuron $N$. The first set, $\mathbf{C}$, consists of the images that carry the \textit{characteristics} that the specialized neuron $N_S$ should respond to.
The second set, $\mathbf{R}$, consists of all the remaining  images that do not. The discernibility score is composed of two score functions. The first one, called \emph{fire rate score}, indicates the tendency of a neuron to fire only for the set $\mathbf{C}$. The second score, called \emph{activation score}, highlights how strong this tendency appears to be. By averaging the two scores, $\mathcal{D}$ indicates both the tendency and
the strength of a
 neuron response for the set $\mathbf{C}$.
The score has to be applied on each neuron under analysis; other than finding the most involved neuron as in Eq. \ref{Eq:disc}, the score can be used to sort the neurons with respect to their sensitivity to $\mathbf{C}$.

We first define the fire rate score $\mathcal{F}$ as the difference in mean neuron activity over the sets $\mathbf{C}$ and $\mathbf{R}$, as in Eq. \ref{eq:Fire_rate}, where $\#\mathbf{C}$ and $\#\mathbf{R}$ represent the cardinalities of the aforementioned sets:

\begin{equation}
\begin{split}
\mathcal{F}(\mathbf{C},\mathbf{R},N)  = ~&~  \left (\frac{1}{\#\mathbf{C}} \sum_{\mathbf{X}\in \mathbf{C}}^{} T( A_{N}(\mathbf{X})) \right)  \\
 & - \>   \left (\frac{1}{\#\mathbf{R}} \sum_{\mathbf{X}\in \mathbf{R}}^{} T( A_{N}(\mathbf{X})) \right),  
\end{split}
\label{eq:Fire_rate}    
\end{equation}

where $\mathbf{X}$ is the input image, $A_{N}$ is the activation of neuron $N$ in a given layer and $T$ is a threshold function here selected to be the Heaviside step function:

\begin{equation}
        T(x)=
\begin{cases}
    1,& \text{if } x > 0\\
    0,              & \text{otherwise}.
\end{cases}
\label{eq:tr}
\end{equation}

In our case, $N$ is a neuron in the fully connected layer preceding the softmax layer. $T$ was set to the Heaviside step function because $A_{N}$ of the probed layer is zero-centered.

The activation score $\mathcal{A}$ is the normalized difference between the mean activations of subsets $C$ and $R$ as defined in Eq.~\ref{eq:act}. The normalizing factors of $\frac{1}{2}$ are due to the use of hyperbolic tangent as activation unit. These will ensure a maximum activation score $\mathcal{A}$  of $1$:

\begin{equation}
\begin{split}
\mathcal{A}(\mathbf{C},\mathbf{R},N)   = &  \frac{1}{2}  \left (\frac{1}{\#\mathbf{C}} \sum_{\mathbf{X}\in \mathbf{C}}^{}  A_{N}(\mathbf{X}) \right) \\
&-  \frac{1}{2}  \left (\frac{1}{\#\mathbf{R}} \sum_{\mathbf{X}\in \mathbf{R}}^{}  A_{N}(\mathbf{X}) \right).
\end{split}
\label{eq:act}
\end{equation}

Finally, discernibility $\mathcal{D}$  is defined as the mean of  the fire rate and activation scores:

\begin{equation}
\mathcal{D}(\mathbf{C},\mathbf{R},N) = \frac{1}{2} \cdot \left( \mathcal{F}(\mathbf{C},\mathbf{R},N) + \mathcal{A}(\mathbf{C},\mathbf{R},N) \right).
\label{eq:discern}
\end{equation}

To investigate the role that the neurons of SOMAnet play in encoding the human figure, we use the discernability measure defined in Sec.~\ref{sec:probe} on two datasets: the synthetic SOMAset and the real RGBD-ID. Given a dataset,
we partition it into two groups, the \emph{localization} $\mathbf{L}$ and the \emph{exploration} $\mathbf{E}$.  The images in $\mathbf{L}$ are used to localize the specialized neurons w.r.t a structural characteristic (as being obese);  in particular, the set $\mathbf{L}$ is manually subdivided in  $\mathbf{C}$ (with images of subjects with that characteristic) and $\mathbf{R}$ (absence of that characteristic), in order to compute the discernability measure  (see Eq.~\ref{eq:discern}) . Subsequently, the images in $\mathbf{E}$ triggering the specialized neurons the most are analyzed, looking for analogies with the images in $\mathbf{C}$ .
In general, we focus on visual characteristics that are present in a sufficient number of samples of a dataset: for SOMAset, we analyze obesity and gender, while for the RGBD-ID we analyse ectomorphism (being long and lean) and a particular kind of clothing, independently on color information.

\begin{figure}[!htb]
  \centering
  \subfigure[Obesity]{\includegraphics[width=0.495\columnwidth]{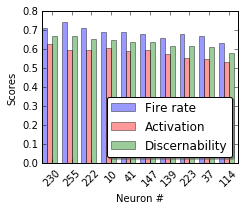}}%
  \hfill
  \subfigure[Gender]{\includegraphics[width=0.495\columnwidth]{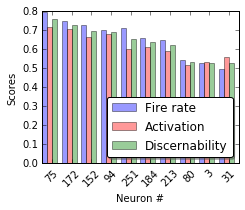}}
  \caption{Top 10 neurons ranked by their discernability score $\mathcal{D}$.
  }
  \label{fig:Female_and_obese}
\end{figure}

In the case of SOMAset, $\mathbf{L}$ contains 64,000 images from 32 randomly selected subjects, 16 female and 16 male, while $\mathbf{E}$ contains 36000 images from the remaining 18 subjects - 9 female and 9 male.
For the obesity trait, $\mathbf{C}$ contains 4,000 images from two obese subjects and $\mathbf{R}$ the remaining 60,000 ones from the other 30 subjects. Using Eq.~\ref{eq:Fire_rate}  and \ref{eq:act} we compute the fire rate $\mathcal{F}$ and the activation $\mathcal{A}$, averaging them to get the discernability score $\mathcal{D}$ (Eq.~\ref{eq:discern}). This process is carried out for each neuron, producing at the end a ranking of the most responsive neurons. Heuristically, we select the top 10 of them, as giving good results when it comes to the analysis of $\mathbf{E}$; their values for $\mathcal{F}$, $\mathcal{A}$ and $\mathcal{D}$ are shown in Fig.~\ref{fig:Female_and_obese}a.
As visible, the ranking shows the neurons reacting in a similar way, and this could mean that they are cooperating together to explain the data, in line with the distributed representation theory~\cite{hinton1986,plate2002,valdez2015distributed}. An automatic selection of the number of neurons required to represent a visual characteristic is still an open topic planned for future research.

\begin{figure}[!htb]
  \centering
  \includegraphics[width=.80\columnwidth]{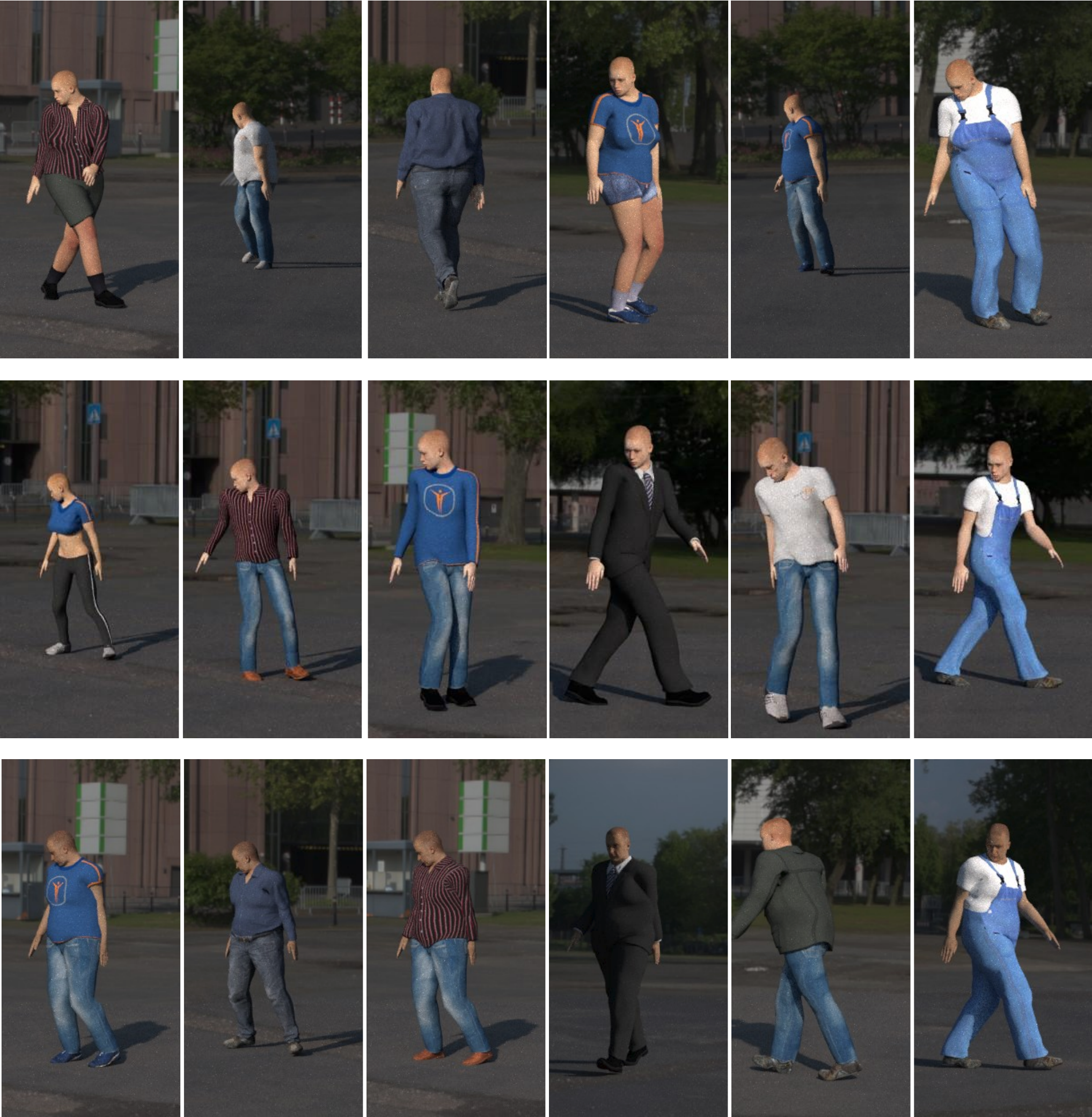}
  \caption{ The first two rows show images used to find specialized neurons: the first row has images of obese subjects ($\in \mathbf{C}$), the second shows subjects without such characteristic ($\in \mathbf{R}$). The third row shows the test images $\in \mathbf{E}$ which responded to the specialized neurons. As visible, all of them portray obese subjects.}
  \label{fig:obese_soma}
\end{figure}

Subsequently, on the set $\mathbf{E}$, we extract those images that cause the network to have as most discerning neurons the same 10 found on the set $\mathbf{L}$. In the majority of the cases, obese subjects pop out. A random sampling of the images is shown in Fig.~\ref{fig:obese_soma}.

In the case of gender, $\mathbf{C}$ contains all the images from $\mathbf{L}$ where the subject is female (32000 elements), and  $\mathbf{R}$ contains the male subjects (the other 32000). Even in this case, the top 10 neurons in terms of discernability score are kept (see Fig.~\ref{fig:Female_and_obese}b). Furthermore, $\mathcal{D}$ shows to decrease more rapidly than for the obesity case; this could mean that the gender trait is more easily detectable, requiring less neurons to focus on it. In fact, on the exploration set $\mathbf{E}$, all the female subjects have been detected correctly.

\begin{figure}[!htb]
  \centering
  \subfigure[\vspace{-0.1cm}Ectomorph]{\includegraphics[width=0.3428\columnwidth]{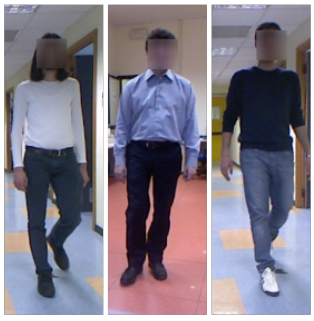}} \vspace{-0.1cm}%
  \hfill \addtocounter{subfigure}{2}
  \subfigure[\vspace{-0.1cm}Long-sleeved]{\includegraphics[width=0.4571\columnwidth]{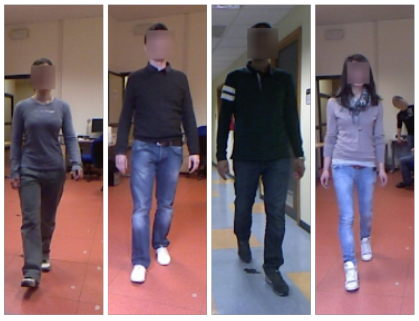}}\vspace{-0.1cm}
  \addtocounter{subfigure}{-3}
  \subfigure[\vspace{-0.1cm} Not  ectomorph]{\includegraphics[width=0.3428\columnwidth]{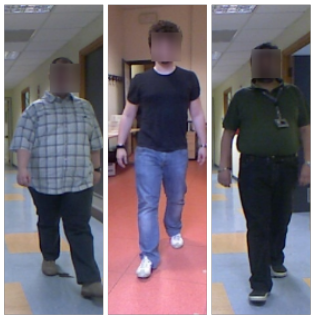}}\vspace{-0.1cm}%
  \hfill  \addtocounter{subfigure}{2}
  \subfigure[\vspace{-0.1cm}Not long-sleeved]{\includegraphics[width=0.4571\columnwidth]{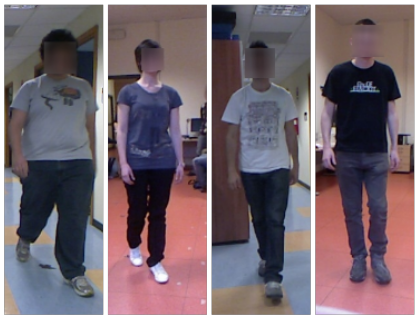}}\vspace{-0.1cm}
  \addtocounter{subfigure}{-3}
  \subfigure[Images triggering \emph{ectomorph} neurons]{\includegraphics[width=0.3428\columnwidth]{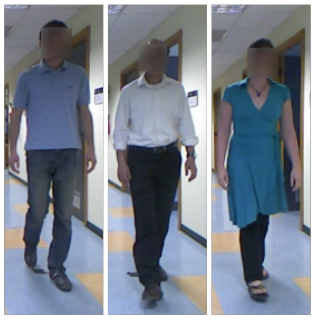}}\vspace{-0.1cm}%
  \hfill \addtocounter{subfigure}{2}
  \subfigure[Images triggering \emph{long sleeve} neurons]{\includegraphics[width=0.4571\columnwidth]{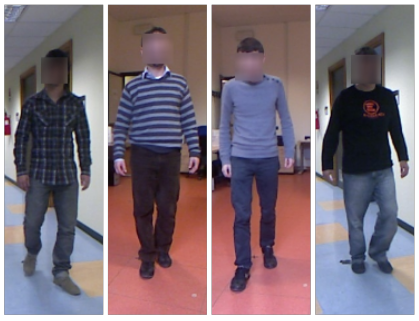}}\vspace{-0.1cm}

  \caption{The first two rows show images used to find specialized neurons: the first row has images of subjects with the visual characteristic ($\in \mathbf{C}$), the second has images that do not have it  ($\in \mathbf{R}$). The third row shows random test images $\in \mathbf{E}$ which responded to the specialized neurons.
  }
  \label{fig:tall_obese}
\end{figure}

Concerning the analysis on the real RGBD-ID dataset, we use SOMAnet, trained on SOMAset and fine-tuned on the RGBD-ID, called here \ts{SOMAnet}{SOMAset+RGBD-ID}. We first find neurons that respond to \emph{ectomorph} (long and lean) subjects and subjects with long-sleeved shirts, respectively.
In particular, the localization set $\mathbf{L}$ is composed of 59 subjects,  and the exploration set $\mathbf{E}$ of the remaining 20; both partitions contain subjects that possess or not the characteristic. In the same way as for the previous experiments, we find the 10 top neurons that respond to the presence of the characteristic (high positive $\mathcal{D}$ ), and subsequently we check those images of the exploration set which trigger the same 10 neurons. Results are shown in Fig.~\ref{fig:tall_obese}.

These experiments show that SOMAnet sees beyond the appearance of the human silhouette, capturing structural aspects, which are capable of boosting the re-id performance.

\subsubsection{SOMAnet + SOMAset}\label{Sec:topexp}
We next analyze the re-id performance when SOMAnet is trained from scratch with SOMAset, and fine-tuned on the training partition of another dataset, whose testing partition is used to calculate the re-id figures.
In this case, we analyze the performance on all the four datasets, comparing against the approaches that, at the time of writing, exhibit the best performance.

For CUHK03, results of the single- and multi-shot modalities are reported in terms of CMC curves and mAP in Table~\ref{Table:CUHK03_SOMA50_multishot}.

Here SOMAnet has been trained from scratch on SOMAset and fine-tuned on CUHK03, labelled
\ts{SOMAnet}{SOMAset+CUHK03}. The resulting classifier is competitive against the state-of-the-art. Concerning the multi-shot modality, the only approach that operates on the CUHK03 dataset is MB-DML, which provides    results just for rank 1 of the CMC curve.

We also report the scores obtained with SOMAnet, trained from scratch on CUHK03 (these are the results reported in Sec.~\ref{Sec:vsDeep}), to show the advantage of bringing in SOMAset into play. 


\begin{table}[!htbp]
\scriptsize
	\centering
	\caption{Analysis of the performance of SOMAnet against other methodologies when trained on SOMAset,  fine-tuned on the training partition of the CUHK03 dataset, and tested on the test partition of the CUHK03 dataset, in both \emph{single-shot} and \emph{multi-shot} modalities.}
	\scalebox{0.96}{
	\begin{tabular}{cccccc}
	\toprule
        \multirow{ 2}{*}{\textbf{Single-shot}} & \multicolumn{5}{c}{\textbf{Rank}} \\
		\cmidrule(r){2-6}
		 \textbf{} & \textbf{1} & \textbf{2}& \textbf{5} & \textbf{10} & \textbf{20} \\
		\midrule
		KISSME \cite{KISSME}                       & 14.17 & 22.30 & 37.46 & 52.20 & 69.38 \\
		FPNN \cite{li2014deepreid}                 & 20.65 & 32.53 & 50.94 & 67.01 & 83.00 \\
		LOMO+XQDA \cite{Liao_2015_CVPR}            & 52.20 & 66.74 & 82.23 & 92.14 & 96.25 \\
		JointRe-id \cite{Ahmed15}                  & 54.74 & 70.04 & 86.50 & 93.88 & 98.10 \\
		LSTM-re-id \cite{LSTM_REID}                & 57.30 & -     & 80.10 & 88.20 & -     \\
		LOMO+MLAPG \cite{Liao_2015_ICCV}           & 57.96 & -     & -     & -     & -     \\
		Ensemble \cite{Paisitkriangkrai_2015_CVPR} & 62.10 & 76.60 & 89.10 & 94.30 & 97.80 \\
		Null space \cite{Zhang_2016_CVPR}          & 62.55 & -     & 90.05 & 94.80 & 98.10 \\
        Personnet \cite{personnet}                 & 64.80 & 73.55 & 89.40 & 94.92 & 98.20 \\
        MB-DML \cite{mbdml}                        & 65.04 & -     & -     & -     & -     \\
        Gated \cite{GATED}                         & 68.10 & -     & 88.10 & 94.60 & -     \\

        DGD-CNN \cite{Xiao_2016_CVPR}     & 72.60 & -     & -     & -     & -     \\
        MTDnet \cite{chen2017multi}       & \textbf{74.68} & - & \textbf{95.99} & \textbf{97.47} & - \\

        \ts{SOMAnet}{CUHK03}                     & 68.90 & 82.10  & 91.00 & 95.60 & 98.30 \\
        \ts{SOMAnet}{SOMAset+CUHK03}             & 72.40  & 81.90 & 92.10 & 95.80 & 98.50   \\
        \midrule
		\multirow{ 2}{*}{\textbf{Multi-shot}} & \multicolumn{5}{c}{\textbf{Rank}} \\
		\cmidrule(r){2-6}
		\textbf{} & \textbf{1} & \textbf{2}& \textbf{5} & \textbf{10} & \textbf{20} \\\midrule
        MB-DML \cite{mbdml}                       & 80.60 & -     & -     & -     & -     \\
        \ts{SOMAnet}{CUHK03}    & {83.60}   & {93.40}   & {97.50}   & {99.20}   & \textbf{99.70}   \\
    \ts{SOMAnet}{SOMAset+CUHK03}            & \textbf{85.90}   & \textbf{94.00}   & \textbf{98.10}   & \textbf{99.30}   & {99.60}   \\
    \bottomrule
    \end{tabular} \label{Table:CUHK03_SOMA50_multishot}
    }
\end{table}

Results on Market-1501 are reported in Table~\ref{TTable:Market-1501TOP} along with the competitive approaches.
We also report here the scores obtained with SOMAnet, trained from scratch on Market-1501.  For the single-shot evaluation \ts{SOMAnet}{SOMAset+Market-1501} provides a mAP  of $47.89\%$.

\begin{table}[!htbp]
\scriptsize
	\centering

	\caption{Analysis of the performance of SOMAnet against other methodologies when trained on SOMAset,  fine-tuned on the training partition of the Market-1501 dataset, and tested on the test partition of the same dataset, in the \emph{single-shot} and \emph{multi-shot} modalities.}
\label{TTable:Market-1501TOP}
\scalebox{0.80}{
\begin{tabular}{cccccccc}
		\toprule
		\multirow{ 2}{*}{\textbf{Single-shot}} & \multicolumn{6}{c}{\textbf{Rank}}  & \multirow{ 2}{*}{\textbf{mAP}} \\
		\cmidrule(r){2-7}
		\textbf{}& \textbf{1} & \textbf{5}& \textbf{10} & \textbf{20} & \textbf{30} & \textbf{50}&  \\
		\midrule
		 BoW \cite{market} & 35.84 & 52.40 & 60.33 & 67.64& 71.88& 75.80 & 14.75\\
		 BoW,LMNN \cite{market} & 34.00 & - & - & - & - & - & 15.66\\
		 BoW,ITML \cite{market} & 38.21 & - & - & - & - & - & 17.05\\
		 Personnet \cite{personnet}    & 37.21 & -       & -       & -       & -       & -       &  18.57 \\
		 SSDAL \cite{deep_atrib} & 39.40 & -       & -       & -       & -       & -       & 19.60\\
 		 BoW,KISSME \cite{market}& 44.42 & 63.90 & 72.18 & 78.95& 82.51& 87.05 & 20.76\\
         SCSP \cite{Chen_2016_CVPR} & 51.90 & -       & -       & -       & -       & -       & 26.35 \\
         Null space \cite{Zhang_2016_CVPR} & 61.02 & -       & -       & -       & -       & -       & 35.68\\
         Gated \cite{GATED}  & 65.88 & -       & -       & -       & -       & -       & 39.55\\
        \ts{SOMAnet}{Market-1501} & {70.28} & {87.53} & {91.69} & {94.57} & {95.64} & {96.85} & {45.05}\\
         \ts{SOMAnet}{SOMAset+Market-1501}       & \textbf{73.87} & \textbf{88.03} & \textbf{92.22} & \textbf{95.07} & \textbf{96.20} & \textbf{97.39} & \textbf{47.89}\\
         \midrule
         \multirow{ 2}{*}{\textbf{Multi-shot}} & \multicolumn{6}{c}{\textbf{Rank}}  & \multirow{ 2}{*}{\textbf{mAP}} \\
		\cmidrule(r){2-7}
		\textbf{}& \textbf{1} & \textbf{5}& \textbf{10} & \textbf{20} & \textbf{30} & \textbf{50}&  \\
		\midrule
		BoW \cite{market} & 44.36 & 60.24 & 66.48 & 73.25& 76.19& 76.69 & 19.42\\
		SSDAL \cite{deep_atrib} & 49.00 & -  & -& -  & -  & -  & 25.80\\
	    LSTM-re-id \cite{LSTM_REID}  & 61.60 & -& - & - & - & - & 35.30\\
		Null space \cite{Zhang_2016_CVPR} & 71.56 & - & - & - & - & - & 46.03\\
		Gated \cite{GATED}  & 76.04 & - & - & - & - & -   & 48.45\\
		\ts{SOMAnet}{Market-1501} & {77.49} & {91.81} & {94.69} & {96.56} & {97.27} & {98.25} & {53.50}\\
		\ts{SOMAnet}{SOMAset+Market-1501}       & \textbf{81.29} & \textbf{92.61} & \textbf{95.31} & \textbf{97.12} & \textbf{97.68} & \textbf{98.43} & \textbf{56.98}\\
    \bottomrule
    \end{tabular}
    }
\end{table}

The third dataset under analysis is RAiD, useful for evaluating the behavior of the SOMA approach when few data are available to fine-tune the network. The dataset has so far just been employed for the multi-shot modality.
The CMC scores on RAiD are reported in Table~\ref{Table:RaidTOP}

\begin{table}[!h]
\scriptsize
\centering
\begin{minipage}[b]{0.3\columnwidth}
	\caption{Analysis of the performance of SOMAnet against other architectures when trained from scratch with the training RAID dataset, and tested on the test partition of the same dataset, in the \emph{multi-shot} modality.} \label{Table:RaidTOP}
\end{minipage}
\scalebox{0.90}{
	\begin{tabular}{cccccc}
		\toprule
		\multirow{ 2}{*}{\textbf{}} & \multicolumn{5}{c}{\textbf{Rank}} \\
		\cmidrule(r){2-6}
		 cam1-cam3& \textbf{1} & \textbf{2}& \textbf{5} & \textbf{10} & \textbf{20} \\
		\midrule
		Double-view \cite{Zhang_2015_ICCV}  & 46.67 & 90.00 & 96.67  & 98.33  & 100.00\\
		NCR on ICT \cite{Das2014}   & 60.00 & 82.00 & 95.00  & 100.00 & 100.00\\
		Multi-view \cite{Zhang_2015_ICCV}   & 61.67 & 91.67 & 96.67  & 100.00 & 100.00\\
		NCR on FT \cite{Das2014}    & 67.00 & 83.00 & 93.00  & 98.00  & 100.00\\
		\ts{SOMAnet}{SOMAset+RAID} & \textbf{69.00} & \textbf{99.00} & \textbf{100.00} & 100.00 & 100.00\\
		\midrule
		\multirow{ 2}{*}{\textbf{}} & \multicolumn{5}{c}{\textbf{Rank}} \\
		\cmidrule(r){2-6}
		 cam1-cam2& \textbf{1} & \textbf{2}& \textbf{5} & \textbf{10} & \textbf{20} \\
		\midrule
		Multi-view \cite{Zhang_2015_ICCV}   & 78.33 & 98.33  & 100.00 & 100.00 & 100.00 \\
		NCR on FT \cite{Das2014}    & 86.00 & 97.00  & 100.00 & 100.00 & 100.00 \\
		Double-view \cite{Zhang_2015_ICCV}  & 88.33 & 100.00 & 100.00 & 100.00 & 100.00 \\
		NCR on ICT \cite{Das2014}   & 89.00 & 98.00  & 100.00 & 100.00 & 100.00 \\
		\ts{SOMAnet}{SOMAset+RAID} & \textbf{95.00} & 100.00 & 100.00 & 100.00 & 100.00\\
		\midrule
		\multirow{ 2}{*}{\textbf{}} & \multicolumn{5}{c}{\textbf{Rank}} \\
		\cmidrule(r){2-6}
		 cam1-cam4& \textbf{1} & \textbf{2}& \textbf{5} & \textbf{10} & \textbf{20} \\
		\midrule
		NCR on ICT \cite{Das2014}   & 66.00 & 84.00  & 94.00  & 100.00 & 100.00 \\
		Multi-view \cite{Zhang_2015_ICCV}   & 66.67 & 98.33  & 100.00 & 100.00 & 100.00 \\
		NCR on FT \cite{Das2014}    & 68.00 & 86.00  & 99.00  & 99.00  & 100.00 \\
		Double-view \cite{Zhang_2015_ICCV}  & 76.67 & 100.00 & 100.00 & 100.00 & 100.00 \\
		\ts{SOMAnet}{SOMAset+RAID} & \textbf{90.00} & 100.00 & 100.00 & 100.00 & 100.00\\
    \bottomrule
    \end{tabular}
}
\end{table}

SOMA saturates CMC performance very soon for all camera combinations,
in several cases starting as early as rank 2. This supports the fact that
fine-tuning SOMAnet on a small dataset worked
appropriately.

Finally, the last dataset we take into account is RGBD-ID. Results on the ``Walking1'' vs ``Walking2'' setting are reported in Table~\ref{Table:RGBd-idTOP}. In order to compare against human performance, we also conducted an experiment with 20 human annotators who were asked to select the top 5 subjects to a given query image (thus producing results for Rank 1 and Rank 5); we report the average performance of the 20 annotators in the same Table.  The annotators complained that the task was tedious, time consuming and challenging, especially when blurred faces were involved. They also reported that their selection was based not only on body shapes but also on detecting common accessories between query and gallery.


\begin{table}
\scriptsize
	\centering
	\begin{minipage}[b]{0.3\columnwidth}
	\caption{Analysis of the performance of SOMAnet against other architectures when trained on SOMAset,  fine-tuned with the training RGBD-ID dataset, and tested on the test partition of  RGBD-ID, in the \emph{multi-shot} modality. Average human performance in Rank 1 and Rank 5 also reported for reference.} \label{Table:RGBd-idTOP}
	\end{minipage}
	\scalebox{0.86}{
	\begin{tabular}{ccccccc}
		\toprule
		\multirow{ 2}{*}{\textbf{}} & \multicolumn{5}{c}{\textbf{Rank}} \\
		\cmidrule(r){2-7}
		 & \textbf{1} &  \textbf{5} & \textbf{10} & \textbf{20} &\textbf{30}&\textbf{50}\\
		\midrule
        RGBD-ID \cite{barbosa2012re}  &12.66 & 43.04 & 53.16 & 84.81 & 96.20 & 100.00 \\
        PDM \cite{PDM}     &17.72 & 36.71 & 40.51 & 59.49 & 77.22 & 91.14  \\
    \ts{SOMAnet}{SOMAset+RGBD-ID}& \textbf{63.29}& \textbf{82.28} & \textbf{88.61} & \textbf{94.94} & \textbf{96.23}  & \textbf{98.73}    \\
    \midrule
    Average Human Performance & 65.00 & 95  & - & - & - & - \\
    \bottomrule
    \end{tabular}}
\end{table}

Notably, the competing approaches work either with the silhouette \cite{PDM}
or with depth images associated to the RGB images (which are not used in the case of SOMA) \cite{barbosa2012re}. As mentioned in Sec.~\ref{Sec:RGBDdataset}, the reason is that people change their clothing between different camera acquisitions. In our case, we do the opposite, discarding depth information while retaining the RGB images only; still the results are well above the state-of-the-art. \ts{SOMAnet}{SOMAset+RGBD-ID} has almost the same rank 1 performance as the average human but SOMAnet is much faster.

\begin{figure}[ht]
  \centering
  \includegraphics[width=0.85\columnwidth]{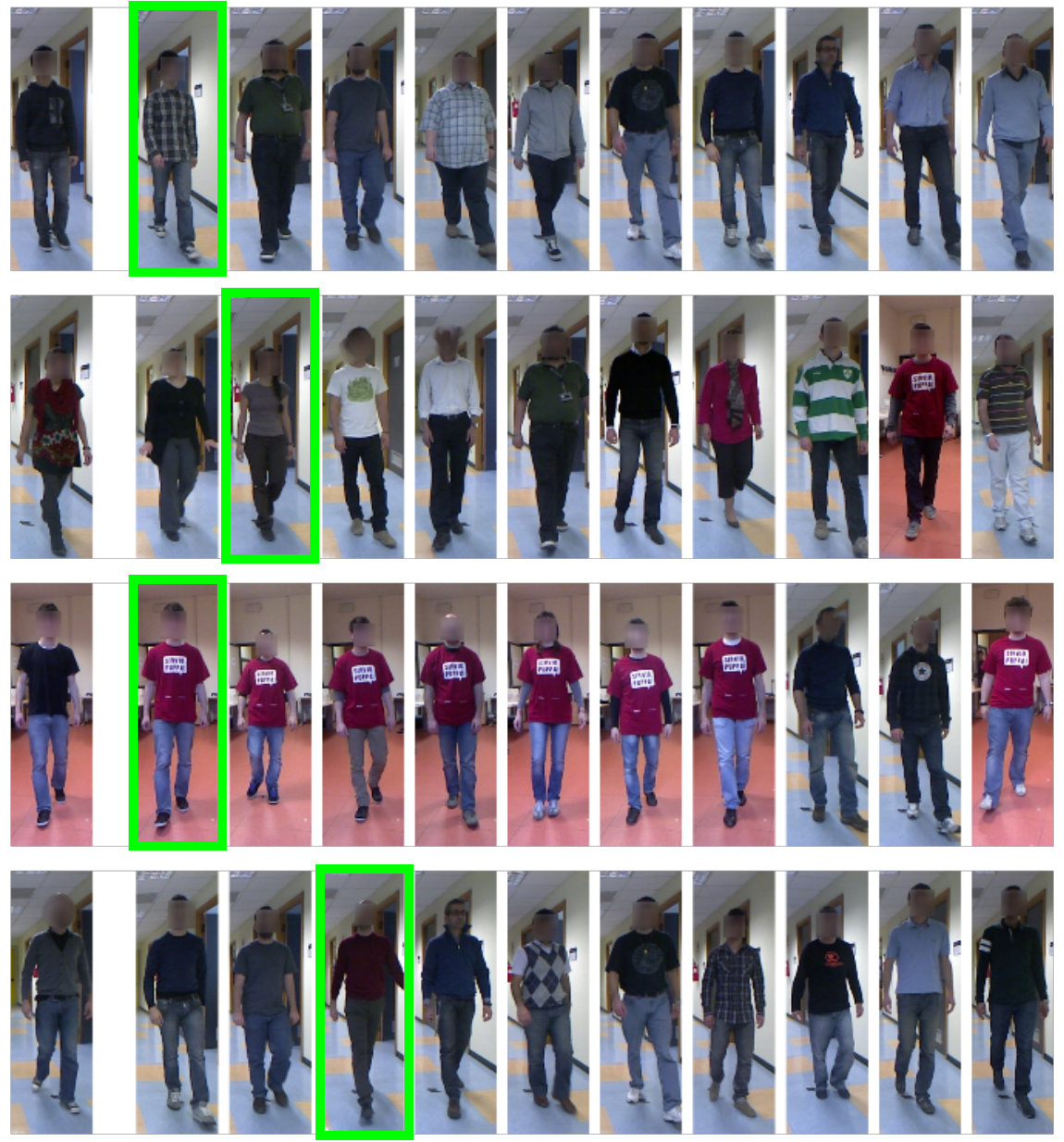}
  \caption{Ranking results of RGBD-ID; probe images are shown in left column. The top 10 ranked gallery images are shown on the right. The ground-truth match is highlighted with a green frame.}
  \label{fig:RGBdidqualitative}
\end{figure}

Fig.~\ref{fig:RGBdidqualitative} illustrates some probe images in the left column and the corresponding ranked gallery images provided by our approach in the rest of the columns, where the correct match is framed in green. As visible, in most of the cases, the correct individual is in the early ranks \emph{even if he/she does not wear the same clothing}. Interestingly, in the second row of Fig.~\ref{fig:RGBdidqualitative}, the probe image of the woman produced two images of women in the top 2 ranks, with another woman in seventh position. In the case of the other probe male subjects, no female subjects appeared in the top ranked gallery images. In the first row, the probe subject is wearing a jacket which is abundant on the belly. Consequently, many of the top ranked gallery images are of endomorph subjects. In contrast, the probe subjects on the third and fourth rows are ectomorph and so are the majority of the retrieved images in the top ranks.

\subsection{Analysis of SOMAset}\label{sec:SOMAset}
This section explores different characteristics of SOMAset, clarifying their role in the re-identification task.
First, we evaluate the importance of training SOMAnet on SOMAset from scratch, \emph{independently of the dataset used to perform fine-tuning and testing}.
To this end, we evaluate the training from scratch with diverse datasets (in addition to SOMAset), choosing different datasets for the fine-tuning  and testing (for example, we train SOMAnet from scratch on CUHK03, fine-tuning and testing on Market-1501).
Second, we evaluate the effect of reducing the number of different subjects and the number of poses.

\subsubsection{Training from scratch on different datasets} \label{Sec:ScratchTraining}
The datasets that are suitable for training deep networks from scratch are CUHK03 and  Market-1501, due to their size (see Sec.~\ref{Sec:SOMAnetEvaluation}).
In particular, we calculate re-id scores (rank 1 of CMC curve and mAP for brevity)  where SOMAnet is trained with CUHK03, fine-tuned and tested with Market-1501, labelled \ts{SOMAnet}{CUHK03+Market-1501}; then we consider the network  trained on SOMAset, fine-tuned and tested on Market-1501, labelled \ts{SOMAnet}{SOMAset+Market-1501}. To show the effect of this cross-dataset learning, we  also present the results where SOMAnet is trained with Market-1501 from scratch and tested on it, labelled  \ts{SOMAnet}{Market-1501} (these latter  are the results already presented in the experiments of Sec.~\ref{Sec:SOMAnetEvaluation}).

We next invert the roles of CUHK03 and Market-1501, that is, CUHK03 is employed as evaluation dataset, giving rise to \ts{SOMAnet}{Market-1501+CUHK03}, \ts{SOMAnet}{SOMAset+CUHK03} and \ts{SOMAnet}{CUHK03}. All of these setups are evaluated in both the single and multi-shot modalities. Results are reported in Table~\ref{Table:Cross_Dataset_Market-1501}
and Table~\ref{Table:Cross_Dataset_CUHK03}, respectively.

\begin{table}[!htbp]
\scriptsize
	\centering
	\caption{Analysis of the role of SOMAset as learning data for the training from scratch step of SOMAnet. For the same testing dataset, Market-1501, different repositories are used for the training from scratch, namely, Market-1501 itself, CUHK03 and SOMAset, respectively.}
	\begin{tabular}{cccc}
	\toprule
	\multicolumn{4}{c}{\textbf{Mean Average Precision}}\\
	\midrule
	\textbf{} & \ts{SOMAnet}{Market1501} &\ts{SOMAnet}{CUHK03+Market1501} & \ts{SOMAnet}{SOMAset+Market1501}  \\
	\cmidrule{2-4}
	Single-shot& 45.05 & 45.97 & \textbf{47.89} \\
	Multi-shot& 53.50  & 54.20 & \textbf{56.98} \\
	\midrule
	\multicolumn{4}{c}{\textbf{Rank 1}}\\
	\midrule
    \textbf{} & \ts{SOMAnet}{Market1501} &\ts{SOMAnet}{CUHK03+Market1501} & \ts{SOMAnet}{SOMAset+Market1501}  \\
	\cmidrule{2-4}
	Single-shot& 70.28 & 73.22 & \textbf{73.87} \\
	Multi-shot& 77.49  & 79.81 & \textbf{81.29} \\
    \bottomrule
    \end{tabular} \label{Table:Cross_Dataset_Market-1501}
\end{table}

\begin{table}[!htbp]
\scriptsize
	\centering
	\caption{Analysis of the role of SOMAset as training data for the training from scratch step of SOMAnet. For the same testing dataset, CUHK03, different datasets are used for the training from scratch, namely, the CUHK03 itself, Market-1501 and SOMAset, respectively.}
	\begin{tabular}{cccc}
	\toprule
	\multicolumn{4}{c}{\textbf{Mean Average Precision}}\\
	\midrule
	\textbf{} & \ts{SOMAnet}{CUHK03} &\ts{SOMAnet}{Market1501+CUHK03} & \ts{SOMAnet}{SOMAset+CUHK03}  \\
	\cmidrule{2-4}
	Single-shot& 73.92 & 73.91 & \textbf{76.65} \\
	Multi-shot& 86.79  & 87.49 & \textbf{88.60} \\
	\midrule
	\multicolumn{4}{c}{\textbf{Rank 1}}\\
	\midrule
	\textbf{} & \ts{SOMAnet}{CUHK03} &\ts{SOMAnet}{Market1501+CUHK03} & \ts{SOMAnet}{SOMAset+CUHK03}  \\
	\cmidrule{2-4}
	Single-shot& 68.90 & 68.90 & \textbf{72.40} \\
	Multi-shot& 83.60  & 84.40 & \textbf{85.90} \\
    \bottomrule
    \end{tabular} \label{Table:Cross_Dataset_CUHK03}
\end{table}

By observing the two tables, some useful facts emerge. First, cross-dataset learning seems to be beneficial in general, except for the single-shot modality when testing on the CUHK03 dataset, where the performance essentially does not change.
Notably, fine-tuning gives better results when it is carried out with Market-1501 on the network trained from scratch on CUHK03, than  vice-versa. This is possibly due to the larger size of Market-1501 w.r.t CUHK03. When SOMAset is used for the training from scratch, the improvement is systematically very significant.

This is an interesting result, since it indicates that, other than being an economic and effective proxy for real data, the SOMA framework appears to produce a nice general optimization of the network, that later can be properly specialized  using the data where the classifier will be applied.

\subsubsection{Changing the number of subjects}
\label{sec:reduce_sub}
In these experiments, we analyze the effect of reducing the number of subjects of SOMAset. We recall here that each subject (that is, a mixture of somatotypes) gives rise to 2000 images ($\text{250 human poses} \;\times \; \text{8 sets of clothes}$).  The original SOMAset has 50 subjects, and we evaluate the effect of having 32, 16 and 8. These numbers have been obtained by \emph{randomly} removing  people from the dataset, repeating the experiments twice. When we go to fewer than 8 subjects (in particular, we tried 4) the training of SOMAnet produces several dead/deactivated neurons.

The evaluation of the reduced SOMAsets is carried out with fine-tuning and testing on the Market-1501 dataset. The results are given in terms of CMC ranks and mAP in Table~\ref{Table:ReducingPeopleMarket1501}.

\begin{table}[!htbp]
\scriptsize
	\centering
	\caption{Analysis of the role of the \emph{size} of SOMAset as training data. Here SOMAset was rendered in original and reduced versions by \emph{changing the number of rendered subjects}. The different versions of SOMAset were fine-tuned with the training partition of the Market-1501 dataset, and tested on the test partition of the same dataset.}
	\scalebox{0.86}{
	\begin{tabular}{cc|cc}
        \toprule
        \multicolumn{4}{c}{\textbf{Single-shot}}\\
        \midrule
		 \textbf{\#Images in Dataset} &\textbf{\#Subjects in SOMAset} & \textbf{Rank 1}& \textbf{mAP}\\
		 100000 &  50 & 73.87 & 47.89\\
		 64000  &  32 & 73.13 & 46.70\\
		 32000  &  16 & 72.12 & 46.23\\
		 16000  &  8  & 71.70 & 45.77\\
		 \midrule
      \multicolumn{4}{c}{\textbf{Multi-shot}}\\
        \midrule
		 \textbf{\#Images in SOMAset} &\textbf{\#Subjects in SOMAset} & \textbf{Rank 1}& \textbf{mAP}\\
		 100000 &  50 & 81.29 & 56.98\\
		 64000  &  32 & 80.70 & 55.46\\
		 32000  &  16 & 80.14 & 55.35\\
		 16000  &  8  & 78.79 & 54.68\\
    \bottomrule
    \end{tabular}\label{Table:ReducingPeopleMarket1501}
    }
\end{table}

As one can expect, adding subjects leads to increased performance. The curious aspect is that the increase is very mild, both in terms of rank 1 and mAP. A roughly linear relation between number of subjects and the performance seems to hold. 

We should highlight two points: Market-1501 has 750 subjects in the testing set, and having just $1\%$ of performance increase does impact substantially the re-identification capabilities (an increase of 7.5 subjects matched correctly in the first rank); secondly, in the deep network literature it is widely known that the role of fine-tuning is absolutely crucial, much more than the role of the training from scratch.

\subsubsection{Changing the number of poses}\label{sec:reduce_pos}

In the final experiment, we investigate the impact of reducing SOMAset by \emph{randomly} removing poses from the rendering protocol. To compare with Sec.~\ref{sec:reduce_sub}, and understand if it is more important to have more poses or more subjects into play, we select a number of poses that result in the same number of images as in the previous study.

 Specifically, we create reduced datasets with 250, 160, 80 and 40 poses, giving rise to 100K, 64K,32K and 16K images, corresponding to what we obtained with 50, 32, 16 and 8 subjects, respectively.

\begin{table}[!htbp]
\scriptsize
	\centering
	\caption{Analysis of the role of the \emph{size} of SOMAset as training data. Here SOMAset was rendered in original and reduced versions by \emph{changing the number of poses}. The different versions of SOMAset were fine-tuned with the training partition of Market-1501, and tested on the test partition of the same dataset.}
	\scalebox{0.86}{
	\begin{tabular}{cc|cc}
        \toprule
        \multicolumn{4}{c}{\textbf{Single-shot}}\\
        \midrule
		 \textbf{\#Images in Dataset} &\textbf{\#Poses in SOMAset} & \textbf{Rank 1}& \textbf{mAP}\\
		 100000 &  250 & 73.87 & 47.89\\
		 64000  &  160 & 72.39 & 46.08\\
		 32000  &  80  & 71.44 & 45.18\\
		 16000  &  40  & 70.19 & 44.58\\
		 \midrule
      \multicolumn{4}{c}{\textbf{Multi-shot}}\\
        \midrule
		 \textbf{\#Images in SOMAset} &\textbf{\#Poses in SOMAset} & \textbf{Rank 1}& \textbf{mAP}\\
		 100000 &  250 & 81.29 & 56.98\\
		 64000  &  160 & 79.16 & 54.90\\
		 32000  &  80  & 78.65 & 53.72\\
		 16000  &  40  & 78.65 & 53.56\\
    \bottomrule
    \end{tabular}\label{Table:ReducingPosesMarket1501}}
\end{table}

The comparison of Tables~\ref{Table:ReducingPeopleMarket1501} and \ref{Table:ReducingPosesMarket1501} indicates that having more subjects than poses is more auspicable, and this is meaningful, since the intraclass variance of a dataset is intuitively higher when having different subjects instead of different poses, in terms of visual variability (consider the rows starting from the second one, since the first row shows the performance of the full SOMAset, which is the same in both tables).

\subsection{Effects of Illumination, poses and camera viewpoints}

It is interesting to attempt to quantitatively assess the individual effect of illumination, poses and camera viewpoints on performance. Then one could determine which variable to prioritize while modeling and rendering a synthetic dataset for re-id. To this end, we have isolated 4 variants of SOMAset with 16000 images: the first consists of a manual selection of 16000 images where the subject appears dark (bad illumination), the second consists of 16000 images in which the number of rendered poses has been reduced following the procedure of Section \ref{sec:reduce_pos} (restricted poses),  the third consists of a manual selection of 16000 images where the subject is seen from the back (bad viewpoint), while the fourth is a balanced random selection of 16000 images called the control group, for comparison. We have repeated the experiment with 4 similar variants of 32000 images in order to see how the dataset size change influences these factors. The results can be seen in Table \ref{Table:factors}.

\begin{table}[!htbp]
\scriptsize
	\centering
	\caption{Comparitve analysis of rendering factors of SOMAset on SOMAnet performance. The effect of a  balanced control group is compared against similarly sized datasets with bad illumination, restricted number of poses and bad camera viewpoints. The experiment was performed for 16000 and 32000 images giving a total of 8 variants of SOMAset. All variants were fine-tuned with the training partition of Market-1501 and tested on the test partition of the same dataset.}
	\scalebox{0.99}{
	\begin{tabular}{c|cc|cc}
        \toprule
        \multicolumn{5}{c}{\textbf{Multi-shot}}\\
        \midrule
       	 \textbf{SOMAset variant} & \textbf{Rank 1}& \textbf{mAP} &
		 \textbf{Rank 1}& \textbf{mAP}\\
		 \midrule
		 Balanced Control Group      & 80.14 & 55.35 & 78.89 & 54.68\\
		 Bad Illumination   & 79.19 & 54.77 & 71.73 & 54.33\\
		 Bad Viewpoint      & 79.13 & 53.84 & 71.56 & 54.56\\
 		 Restricted Poses   & 78.65 & 53.72 & 70.19 & 53.56\\

    \midrule
     & \multicolumn{2}{c|}{32000 Images} & \multicolumn{2}{c|}{16000 Images} \\
    \cline{2-5}
    \end{tabular}\label{Table:factors}}
\end{table}

As one would expect, the Balanced Control Group performs best across both dataset sizes. Looking at Rank 1 performance, in both dataset sizes, the most degrading factor compared to Balanced Control Group performance is restricting the number of poses, followed by bad viewpoint and bad illumination. The mAP performance generally follows the same pattern, except for the case of bad viewpoint where, paradoxically, mAP performance drops when going from 16000 to 32000 images. The Rank 1 difference between the Balanced Control Group and the 'degraded' variants at 16000 images is over 7\% while the equivalent figure for 32000 images is less than 2\%; this is likely to be due to the fact that overfitting to a 'degraded' dataset is easier the smaller its size is.

\section{Conclusions}
\label{Conclusions}
Synthetic training data can greatly help to initialise deep networks. Tasks such as re-identification should not be faced exclusively by siamese architectures; instead, single-path networks can be employed as successful feature extractors. A by-product is that these networks can be easily probed, investigating the semantics being captured by the neurons.

In this work, we find that such networks can see beyond apparel, capturing structural aspects of the human body, such as their somatotype. This can be fully exploited with an appropriate dataset; in this respect, we introduce, for the first time in the re-identification field, the strategy of using synthetic data as proxy for real data. In particular, having synthetic datasets for training a network from scratch  seems to be a very effective manoeuvre, producing successive fine-tuned architectures with a very high recognition rate. The proposed inception-based network, SOMAnet, trained on the synthetic dataset SOMAset\footnote{SOMAset will be released with a open source license to enable further developments in re-identification.} can match people even if they change apparel between camera acquisitions.

Various future directions are intriguing and promising. First, the nature of the synthetic dataset needs to be explored under different respects: an obvious question is, what is the behavior of the network when the number of subjects contained in the dataset tends to infinity. Specifically, we show a somewhat linear increase in performance with respect to the addition of diverse subjects. Certainly, at a given point, a plateau should be reached, and finding this point is a key open issue.

Another question regards the importance of the background in the images: to bound the degree of freedom of our analysis, we decided to place our synthetic pedestrians in a single scene that, even if arbitrarily large, does not offer the variability contained in other datasets. Our intuition is that having a fixed background forces the network to focus on the foreground objects. At the same time, a single scene may help the network in understanding differences among individuals, acting as a frame of reference to capture, for example, different sizes among individuals. In a preliminary experiment, not reported here intentionally, we omit the background leaving a grey homogeneous flat area behind the subjects. Results in recognition are definitely worse, but we did not investigate this point further.
The importance of having realistic images is another question that we would like to explore. As already mentioned, the usual re-identification setup produces individuals at a certain low resolution, so that fine details such as the face cannot be processed. It could be nice to have an advanced re-id setting, where high-resolution cameras are employed, collecting high frequency cues. In that case it would be reasonable to expect a difference in recognition rates, depending on the realism of the training data.

Finally, a wider, conceptual question pops out: with such a framework, capable of understanding bodily cues of human beings, going beyond the mere appearance of the outfit, is it still reasonable to talk about re-identification, or does it make more sense to call for non-collaborative person recognition at a distance? In that case, a brand new biometric field is opening up.

{
\bibliographystyle{IEEEtran}
\bibliography{egbib}
}

\end{document}